\documentclass[10pt,twocolumn,letterpaper]{article}

\usepackage[pagenumbers]{cvpr} 
\usepackage{graphicx}
\usepackage{amsmath}
\usepackage{amssymb}
\usepackage{booktabs}

\usepackage[pagebackref,breaklinks,colorlinks]{hyperref}

\usepackage[capitalize]{cleveref}
\crefname{section}{Sec.}{Secs.}
\Crefname{section}{Section}{Sections}
\Crefname{table}{Table}{Tables}
\crefname{table}{Tab.}{Tabs.}

\def\cvprPaperID{*****} 
\def\confName{CVPR}
\def\confYear{2022}

\begin{document}

\title{Random Forest Regression for Continuous Affect Using Facial Action Units}

\author{Saurabh Hinduja\\
University of Pittsburgh\\
4200 Fifth Ave, Pittsburgh, PA 15260\\
{\tt\small sah273@pitt.edu}
\and
Shaun Canavan\\
University of South Florida\\
4202 E Fowler Ave, Tampa, FL 33620\\
{\tt\small scanavan@usf.edu}
\and
Liza Jivnani\\
University of South Florida\\
10811, mc Kinley dr, Tampa, FL 33612\\
{\tt\small ljivnani@usf.edu}
\and
Sk Rahatul Jannat \\
University of South Florida\\
4202 E Fowler Ave, Tampa, FL 33620\\
{\tt\small jannat@usf.edu}
\and
V Sri Chakra Kumar\\
Cornell University\\
2 E Loop Rd, New York, NY 10044\\
{\tt\small vk386@cornell.edu}
}

\maketitle

\begin{abstract}
In this paper we describe our approach to the arousal and valence track of the 3rd Workshop and Competition on Affective Behavior Analysis in-the-wild (ABAW). We extracted facial features using OpenFace and used them to train a multiple output random forest regressor. Our approach performed comparable to the baseline approach on valence and outperformed in on arousal.
\end{abstract}


\section{Introduction}
\label{sec:intro}

It has been shown that use hand crafted features to train machine learning models perform significantly  well for affect recognition. Hinduja et al \cite{hinduja2019fusion} showed that fusing hand crafted features, for empathy prediction, worked well using multiple classifier types. Facial action units \cite{FACS} and gaze vectors have been shown to be very useful. Fabinao et al \cite{fabiano2020gaze} used gaze to classify the risk of autism. Srivastava et al \cite{srivastava2020recognizing} used facial action units and gaze vectors for recognizing affect in the wild. Motivated by these methods, we adopt a similar approach to the arousal and valence track of the 3rd Workshop and Competition on Affective Behavior Analysis in-the-wild (ABAW) \cite{kollias2022abaw, kollias2021analysing, kollias2020analysing, kollias2021distribution, kollias2021affect, kollias2019expression, kollias2019face, kollias2019deep, zafeiriou2017aff}. We propose to use hand-crafted features from the cropped frames in the dataset to train a random forest regressor to predict valence and arousal.
\section{Method}
\label{sec:Method}

Our proposed method for detecting frame level valence and arousal uses hand crafted features to train a multiple output random forest regressor. The hand crafted features we selected were action unit occurrence and intensity, the gaze vector, gaze angles, and head pose. To extract these features we used the publicly available Openface \cite{baltrusaitis2018openface}. We did not run open face on the raw videos but on the cropped frames provided. Due to the complexity and the volume of data it took us a long time to pre-process and prepare the hand crafted features for training.

Given the extracted features, we first created a vector of the hand crafted features $E = [l_x,l_y,l_z,r_x,r_y,r_z,avg_x,avg_y,t_x,t_y, t_z,o_x,o_y,o_z,AU_{I1}, \allowbreak AU_{I2},\dots, AU_{I17}, AU_{O1}, AU_{O2},\dots,AU_{O17}]$ where $l_{\{x,~y,~z\}}$ and $r_{\{x,~y,~z\}}$ are the (x, y, z) coordinates of the left and right eye gaze direction vector, respectively. $t_{\{x,~y,~z\}}$ and $o_{\{x,~y,~z\}}$ are the (x, y, z) coordinates of the head pose translation and orientation vectors, respectively. $AU_{Ii}$ and $AU_{Oi}$ are the AU intensity and AU occurrence respectively. Next, we use the this vector to train a 250 trees random forest with multiple output regressor. Using a multiple output regressor helps to use a method not suitable for multiple output to predict multiple outputs. It does by fitting one regressor per output. 
\section{Experiments and Results}
\label{sec:exp}
\begin{table*}
    \centering
        \caption{Comparison of Valence-Arousal Challenge on Validation Set }
 \begin{tabular}{|c|c|c|c|}
 \hline
 Method & CCC-Valence & CCC-Arousal & PVA \\
 \hline
 Baseline(ResNet50) \cite{kollias2022abaw} & 0.31  & 0.17  & 0.24 \\
 \hline
 Ours  & 0.26  & 0.19  & 0.225 \\
 \hline
 \end{tabular}%

    \label{tab:ValidationResults}
\end{table*}
To compare with the baseline, our proposed approach was evaluated using concordance correlation coefficient (CCC) \cite{liao2000note}. Through evaluating the validation set of the challenge data, using the proposed approach, we achieved a CCC score of 0.26 and 0.19 for valence and arousal, respectively. As can be seen in Table \ref{tab:ValidationResults}, while our methods does not beat the baseline for valence, it does outperform it for arousal. The average CC score across both arousal and valence was 0.225, which is comparable to the baseline approach.
\section{Conclusion}
\label{sec:Con}
In this paper, we have shown that hand-crafted features such as actions units, and gaze are a promising approach to predicting continuous affect (i.e., arousal and valence). The proposed approach was comparable to the baseline on valence and outperformed in on arousal. Future work will include the fusion of hand-crafted and deep features using more recent works such as visual transformers.

{\small
\bibliographystyle{ieee_fullname}
\bibliography{egbib}
}

\end{document}